\documentclass[10pt,twocolumn,letterpaper]{article}

\usepackage[accsupp]{axessibility}
\usepackage{cvpr}  
\usepackage{overpic}
\usepackage{bm}
\usepackage{amssymb}
\usepackage{amsmath}
\usepackage{booktabs}
\usepackage{multirow}
\usepackage{marvosym}
\usepackage{makecell} 
\usepackage{color}
\usepackage{xcolor}
\usepackage{colortbl}
\usepackage{pifont}
\usepackage{listings}
\newcommand{\cmark}{\ding{51}}
\newcommand{\xmark}{\ding{55}}
\newcommand{\CC}[1]{\cellcolor{gray!#1}}
\definecolor{shadecolor}{RGB}{150,150,150}
\definecolor{Gray}{gray}{0.5}
\definecolor{Highlight}{HTML}{39b54a}
\definecolor{Highlight2}{HTML}{8FAADC}
\usepackage{soul}
\usepackage{lipsum}
\definecolor{frenchblue}{rgb}{0.0, 0.45, 0.73}

\definecolor{cvprblue}{rgb}{0.21,0.49,0.74}
\usepackage[pagebackref,breaklinks,colorlinks,citecolor=cvprblue]{hyperref}

\def\paperID{****} 
\def\confName{CVPR}
\def\confYear{2024}

\title{Parameter Efficient Fine-tuning via Cross Block Orchestration for Segment Anything Model}

\begin{document}

\author{Zelin Peng$^{1,\dag}$, Zhengqin Xu$^{1,\dag}$, Zhilin Zeng$^{1}$, Lingxi Xie$^{2}$,  Qi Tian$^{2}$, and Wei Shen$^{1{(\textrm{\Letter})}}$\\
	$^1$MoE Key Lab of Artificial Intelligence, AI Institute, Shanghai Jiao Tong University\\  $^2$Huawei Inc.\\
	{\tt\small 
		\{zelin.peng, fate311, bernardeschi, wei.shen\}@sjtu.edu.cn;}  \\ {\tt\small 198808xc@gmail.com; tian.qi1@huawei.com}
}

\maketitle
\newcommand\blfootnote[1]{%
\begingroup 
\renewcommand\thefootnote{}\footnote{#1}%
\addtocounter{footnote}{-1}%
\endgroup 
}
\blfootnote{$^{\textrm{\Letter}}$ Corresponding Author: \texttt{wei.shen@sjtu.edu.cn}}
\blfootnote{$^\dag$ Indicates equal contribution.}
\begin{abstract}

Parameter-efficient fine-tuning (PEFT) is an effective methodology to unleash the potential of large foundation models in novel scenarios with limited training data. In the computer vision community, PEFT has shown effectiveness in image classification, but little research has studied its ability for image segmentation. Fine-tuning segmentation models usually requires a heavier adjustment of parameters to align the proper projection directions in the parameter space for new scenarios. This raises a challenge to existing PEFT algorithms, as they often inject a limited number of individual parameters into each block, which prevents substantial adjustment of the projection direction of the parameter space due to the limitation of Hidden Markov Chain along blocks.  
In this paper, we equip PEFT with a cross-block orchestration mechanism to enable the adaptation of the Segment Anything Model (SAM) to various downstream scenarios.
We introduce a novel inter-block communication module, which integrates a learnable relation matrix to facilitate communication among different coefficient sets of each PEFT block’s parameter space. Moreover, we propose an intra-block enhancement module, which introduces a linear projection head whose weights are generated from a hyper-complex layer, further enhancing the impact of the adjustment of projection directions on the entire parameter space. Extensive experiments on diverse benchmarks demonstrate that our proposed approach consistently improves the segmentation performance significantly on novel scenarios with only around \textbf{1K} additional parameters.

\end{abstract}    
\section{Introduction}
\label{sec:intro}

A notable recent development in AI community is large foundation models~\cite{LM_GPT3_2020_NIPS,FM_SAM_2023_ICCV}, which have made an increasing impact widely across various domains, e.g., natural language processing (NLP)~\cite{NLP_bert_2019_naacl} and computer vision~\cite{CV_vit_2020_arxiv}. 
Despite their surprising zero-shot performance, fine-tuning still remains a crucial step for unleashing their potential in novel scenarios~\cite{zs_SAM_2023_arxiv,SAM_PEFT_MedSAM_arxiv_2023}. To mitigate the extensive fine-tuning costs associated with a large number of pre-trained parameters, parameter efficient fine-tuning (PEFT)~\cite{PEFT_LORA_2022_ICLR,PEFT_Adapter_2019_ICML,PEFT_adaptformer_2022_nips}, which involves tuning a small subset of parameters while maintaining the vast majority frozen, has attracted increasing attentions in various fields.


\begin{figure}[t]
  \centering
  \small
  \begin{overpic}[width=1.0\linewidth]{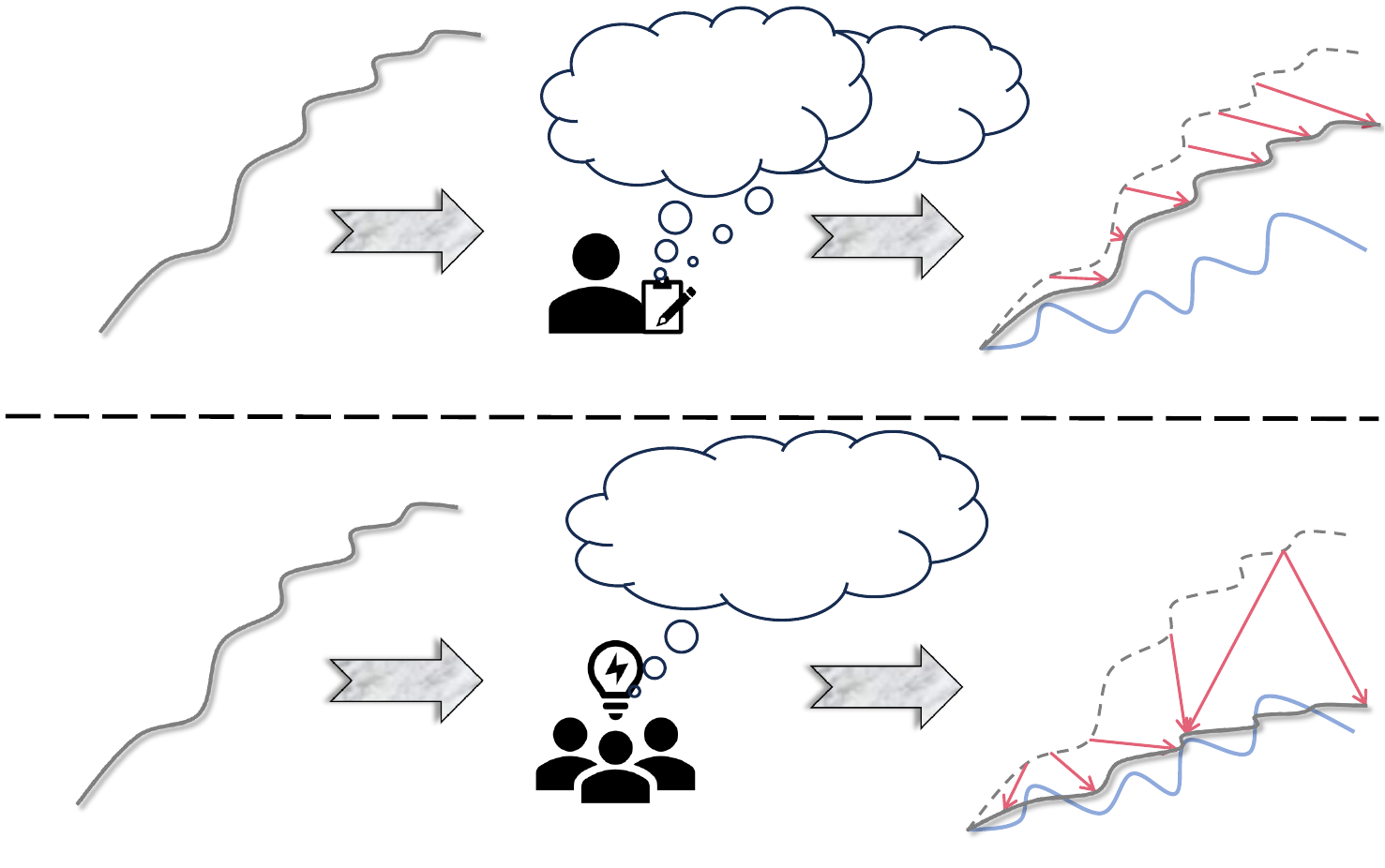}

  \put(0.0,47){(a)}
  \put(0.0,12){(b)}

  \put(0.3,27){SAM's Parameter}
  \put(7.6,22){Space}

  \put(0.3,60){SAM's Parameter}
  \put(7.6,55){Space}
  \put(22.3,-0.5){Projection direction adjustment}
  \put(22.3,35.0){Projection direction adjustment}

  \put(42.3,56.0){\footnotesize{Each block is}}
  \put(43.2,52.0){\footnotesize{\textbf{individual}}}
  \put(64.2,55.0){\tiny{\textbf{HMC}}}
  
  \put(48.1,26.2){\footnotesize{Cross-block}}
  \put(44.7,22.2){\footnotesize{\textbf{Orchestration}}}

  \put(77.3,-0.5){\color{Highlight2}{New Scenarios}}
  \put(77.3,35.0){\color{Highlight2}{New Scenarios}}
  
  \end{overpic}
  \caption{\textbf{Comparison between traditional PEFT
paradigms and our proposed SAM-COBOT.} (a) Traditional methods typically adjust the projection direction of each layer in SAM's parameter space individually, which is limited by the Hidden Markov Chain (HMC). This often leads to relatively minor adjustments. (b) In contrast, our SAM-COBOT approach enhances PEFT with cross-block orchestration, enabling more effective and large adjustments of the projection directions.}
  \label{fig_cof1}
\end{figure}
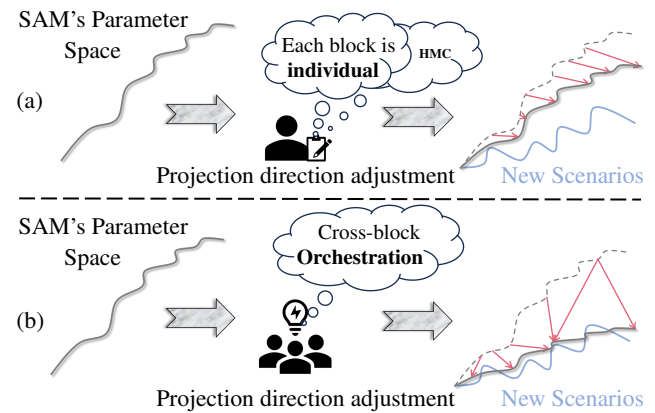

In the field of computer vision, the majority of leading PEFT methodologies have focused on image classification tasks~\cite{PEFT_FacT_2023_AAAI,PEFT_SNF_2023_CVPR,PEFT_VPT_2022_ECCV}. These approaches demonstrate that large classification models can be fine-tuned by injecting a small number of parameters. As evidence, their performance can match or even surpass that of vanilla full fine-tuning~\cite{PEFT_VPT_2022_ECCV}. Influenced by the success of the existing PEFT methods, one can directly apply them in fine-tuning segmentation models~\cite{SAM_PEFT_MedSAM_arxiv_2023,SonarSAM_arxiv_2023}, e.g., Segment Anything Model (SAM)~\cite{FM_SAM_2023_ICCV}. However, fine-tuning segmentation models often necessitates
a heavier adjustment of parameters as the output space for
segmentation is often much larger and more varied compared to classification, which poses a challenge for current PEFT methods that are limited by the Hidden Markov Chain (HMC) along layers~\cite{CVIB_2020_nips,RIB_2021_nips}. The memoryless nature of HMC implies that each layer's state is influenced only by its adjacent layers, thus readily leading to minor adjustments of the projection directions in the entire parameter space for new scenarios, as shown in Fig.~\ref{fig_cof1}.




In this paper, we equip PEFT with a \textbf{C}r\textbf{O}ss-\textbf{B}Lock \textbf{O}rches\textbf{T}ration mechanism to enable the adaptation of SAM to various downstream scenarios, dubbed as SAM-COBOT. The goal of SAM-COBOT is to explicitly integrate cross-block orchestration to enhance the flexibility and reliability of adjusting projection directions. Specifically, we first propose an inter-block communication (IBC) module, which introduces a learnable relation matrix to capture interdependence and facilitate communication among different blocks. The communication is realized by adjusting the coefficient set of each PEFT block's parameter space. We treat all the coefficient sets of PEFT block's parameter space as a tensor, and use the learnable relation matrix to capture cross-slice information for adjusting each coefficient set. Consequently, IBC allows projection directions to influence each other in the entire parameter space. Subsequently, we introduce an intra-block enhancement (IBE) module, which includes a linear project head whose weights are generated from a hyper-complex layer, to ensure that any coordinated adjustments made to the projection directions achieve a greater impact on the entire parameter space. 

Extensive experiments show that the proposed SAM-COBOT can be easily plugged-and-play and consistently improve various PEFT paradigms, e.g., LoRA~\cite{PEFT_LORA_2022_ICLR} and Adaptformer~\cite{PEFT_adaptformer_2022_nips} by a large margin across three prevalent scenarios in computer vision, including natural image segmentation, remote sensing image segmentation, and medical image segmentation. Additionally, SAM-COBOT only needs to introduce around \textbf{1K} parameters (using ViT-Base~\cite{CV_vit_2020_arxiv} as the backbone) while achieving superior segmentation performance.

\section{Related Work}

\noindent\textbf{Parameter Efficient Fine-tuning.}
The objective of parameter-efficient fine-tuning (PEFT) is to utilize the weights from a pre-trained network and tune them for the downstream task by introducing a minimal number of trainable parameters. Many PEFT methods~\cite{PEFT_adapterfusion_2020_arxiv,PEFT_adapterhub_2020_arxiv,PEFT_SNF_2023_CVPR,PEFT_LRA_2023_CVPR} (in deep learning era) are believed to be derived from Adapter~\cite{PEFT_adapter_2017_nips} which introduces a few modules into the pre-trained network. After that, numerous efforts have been devoted to improving the pipeline. For example, some tailored modules like Adapterformer~\cite{PEFT_adaptformer_2022_nips} and RepAdapter~\cite{PEFT_Repadapter_2023_arxiv} are designed for different vision tasks, e.g., classification. Recently, since LoRA~\cite{PEFT_LORA_2022_ICLR}, which replaces additional modules by introducing low-rank matrices to alter the initial parameter spaces, attained growing attention, some other works focusing on automating matrix engineering~\cite{PEFT_Adalora_2023_ICLR, PEFT_FacT_2023_AAAI} to boost the LoRA structure. Apart from the above, some studies also attempt to break new ground, e.g., add extra parameters as prompts along with the inputs~\cite{PEFT_VPT_2022_ECCV}, scales and shifts features after each transformer block~\cite{PEFT_SSF_2022_NIPS}, fine-tune the bias terms in each layer~\cite{PEFT_bitfit_2021_arxiv}, to name a few.

Considering the increasing attention of the large foundation model, i.e., SAM, this paper studies how to boost existing PEFT techniques for fine-tuning SAM~\cite{FM_SAM_2023_ICCV}, from a fresh viewpoint: cross-block orchestration.

\noindent\textbf{Cross-block Orchestration.}
As there exist complementary learning patterns among different blocks (i.e., the shallow block features preserve more details while the deeper one captures more semantics), cross-block orchestration becomes a crucial component of recent state-of-the-art visual recognition algorithms~\cite{PEFT_Cxl_2019_CVPR,PEFT_CSCA_2021_ICCV,seg_segformer_nips_2021,seg_SETR_ICCV_2021}. For example, CLRNet~\cite{CLM_clrnet_2022_CVPR} presented a cross-block refinement module to fully utilize both high-level and low-level features. Zhang~\emph{et al.}~\cite{CLM_sr_2021_ICCV} introduced several self-regulation losses to fully understand detailed features and visual contexts. Chen~\emph{et al.}~\cite{CLM_ada_2019_MM} proposed an adaptive cross-block correlation to recognize the style of visual arts.

This paper also refers to cross-block orchestration. The key differences are that (1) We address the restraint of the Hidden Markov Chain by orchestrating in the parameter space, as opposed to the traditional feature space and (2) for the first time, we introduce a hyper-complex layer~\cite{quaternion_ICLR_2019} to facilitate approaching proper projection directions.

\section{Preliminaries}
\label{sec.3}

\subsection{Hyper-complex Number}

In mathematics, hyper-complex number is a traditional term for an element of a finite-dimensional unital algebra over the field of real numbers. Its elements are generated with real number coefficients $(a_0, \cdots,a_n)$ for a basis $\{ 1, j_1, \cdots, j_n \}$. A $n$-dimensional hyper-complex number is defined in a $n$-dimensional space as:
\begin{equation}
    h = a_01+ a_1{j_1} + a_2{j_2} + \cdots + a_{n-1}{j_{n-1}}.
\end{equation}
In the hyper-complex number, $a_0$ is the real part, $ a_1{j_1} + a_2{j_2} + \cdots + a_{n-1}{j_{n-1}} $ is the imaginary part. For simplicity, we consider $n = 4$ in this work. The imaginary part of a $4$-dimensional hyper-complex number satisfies:
\begin{equation}
    j^2_1 = j^2_2 = j^2_3 = j_1j_2j_3 = -1.
\end{equation}
The geometric interpretation of $j_1$, $j_2$, and $j_3$ can be understood as rotations in $\mathbb{R}^{3}$ space: the $j_1$ rotation represents the rotation from the X-axis to the Y-axis in the plane intersecting both axes, the $j_2$ rotation represents the rotation from the Z-axis to the X-axis in the plane intersecting both axes, and finally, the $j_3$ rotation represents a rotation from Y-axis to Z-axis in a plane that intersects both axes. The negative counterparts $-j_1$, $-j_2$, and $-j_3$ represent reverse rotations of their respective positive counterparts. It is worth noting that, unlike the real and complex numbers, multiplication of the imaginary part of the 4-dimensional hyper-complex number is not commutative, for example, $j_1j_2 = j_3$ and $j_2j_1 = -j_3$.

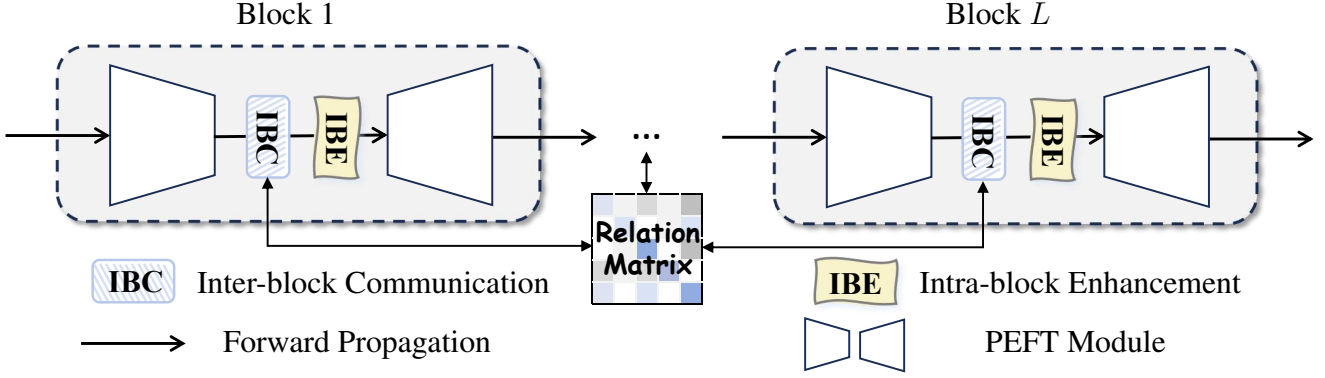
\begin{figure*}[t]
  \centering
  \small
  \begin{overpic}[width=1.0\linewidth]{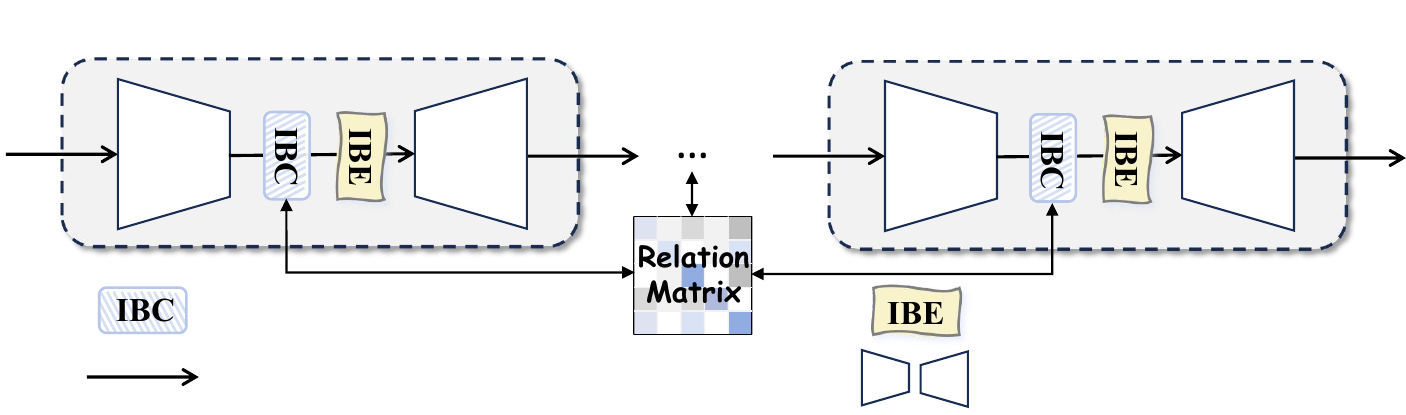}
   \put(15,6.5){\large{Inter-block Communication}}
   \put(17.0,2){\large{Forward Propagation}}
   \put(70,6.5){\large{Intra-block Enhancement}}
   \put(75,2){\large{PEFT Module}}
   \put(18,27){\large{Block 1}}
   \put(72,27){\large{Block $L$}}

  \end{overpic}
  \vspace{-15pt}
  \caption{\textbf{A schematic representation of SAM-COBOT.} In the SAM-COBOT framework, we integrate an inter-block communication module followed by an intra-block enhancement module in each PEFT block.}
  \label{fig_big_picture}
\end{figure*}

\subsection{Segment Anything Model (SAM)}

Segment Anything Model (SAM)~\cite{FM_SAM_2023_ICCV} mainly consists of an image encoder characterized by a vast parameter set, followed by a lightweight mask decoder. The image encoder is structured with $L$ sequential transformer layers. Besides, SAM also incorporates a dedicated prompt encoder, which adaptively handles both dense (mask-based) and sparse (box or point-based) prompts.

\subsection{Parameter Efficient Fine-tuning (PEFT)}
\label{sec.3.2}
\noindent\textbf{Problem Formulation.} Given a large foundation model $\mathcal{F}$, e.g., SAM~\cite{FM_SAM_2023_ICCV}, the goal of PEFT is to fine-tune $\mathcal{F}(\mathbf{X}; \bm{\omega})$ to enable the foundation model to adapt to a new downstream task, where $\mathbf{X}$ is an input image from a dataset of the new task and $\bm{\omega} \in \bm{\Omega} \backslash \bm{\Omega}_l$ denotes the parameters of PEFT modules, which are trainable, while $\bm{\Omega}_l$ refers to the parameter set of $\mathcal{F}$, which are often frozen. Accordingly, the objective function of PEFT is formulated as:
\begin{equation}
    \bm{\omega}^\ast=\arg\min_{\bm{\omega}}\mathcal{L}(\mathbf{X}, {\mathbf{Y}}),
\end{equation}
where $\mathbf{Y}$ is a full dense label map. For segmentation tasks, the loss function $\mathcal{L}$ is commonly selected as a cross-entropy loss. We here briefly review two representative PEFT methods in SAM's adaptation, i.e., Adaptformer~\cite{PEFT_adaptformer_2022_nips} and
LoRA~\cite{PEFT_LORA_2022_ICLR}, since our method is based on them.

\noindent\textbf{Adapterformer.} Adapterformer~\cite{PEFT_adaptformer_2022_nips} introduces a parallel learnable branch for the MLP module in each transformer layer. This branch is primarily composed of a down-projection layer characterized by parameters $\bm{\omega}_{down} \in \mathbb{R}^{D \times V}$ and an up-projection layer represented by parameters  $\bm{\omega}_{up} \in \mathbb{R}^{V \times K}$. Here, the hidden dimension $H$ is significantly smaller than the minimum of $D$ and $K$, i.e., $H \ll \min(D, K)$. The value of $H$ determines the size of the newly introduced parameter space.

\noindent\textbf{LoRA.} LoRA~\cite{PEFT_LORA_2022_ICLR} introduced a learnable low-rank matrix $\bm{\omega}$ that works in parallel to the original weight matrix $\bm{\omega} \in \mathbb{R}^{D \times K}$, which is frequently associated with the parameter matrix in the multi-head self-attention module of each transformer layer. $\bm{\omega}$ is derived by a QR decomposition, denoted as $\bm{\omega} = \bm{\beta}\bm{\alpha}$, where $\bm{\beta} \in \mathbb{R}^{D \times V}$, $\bm{\alpha} \in \mathbb{R}^{V \times K}$.

\noindent\textbf{Drawbacks.} Although existing PEFT methods can be directly integrated with SAM, we notice an opening question in this intuitive solution. Fine-tuning segmentation models often necessitates a heavier adjustment of parameters to align projection directions in the parameter space for new scenarios compared to classification models. However, these PEFT methods introduce only a limited number of parameters in each layer, which can only make relatively small adjustments of projection directions due to the limitation of Hidden Markov Chain (HMC) along SAM's layers. Although some methods, e.g., LST~\cite{LST_2022_NIPS}, seem to bypass this issue by integrating a learnable side adapter, the updating of each layer in the side adapter is also limited to interactions with its adjacent layers. Consequently, the limitations of HMC still exist. In contrast, we devise a plug-and-play method which directly mitigates the constraint of the HMC for existing PEFT methods while introducing nearly zero training efforts.





\section{Methodology}

\subsection{Overview of SAM-COBOT}

Fig.~\ref{fig_big_picture} illustrates the proposed SAM-COBOT framework, which equips PEFT with cross-block orchestration. Each PEFT block's parameter space comes from three parts, (1) PEFT module, (2) Inter-block communication module, and (3) Intra-block enhancement module.

\subsection{Inter-block Communication}
\label{csg}

\noindent\textbf{Coefficient Set Generation.} Following previous studies~\cite{PEFT_Adalora_2023_ICLR,SAM_SAM-PARSER_arxiv_2023}, the parameter space of a PEFT block $\bm{\omega}$ can be decomposed into a base set and a coefficient set. Considering the significantly larger number of base parameters compared to those of coefficients, we opt to facilitate inter-block communication through the coefficient set associated with each block.
To this end, we first introduce a learnable diagonal matrix $\mathbf{\Lambda}$, which is defined as follows:
\begin{equation}
    \mathbf{\Lambda} = \begin{bmatrix}
        \lambda_1 & \cdots & 0  \\
        \vdots & \ddots &\vdots  \\
        0 &  \cdots & \lambda_V
    \end{bmatrix} \in \mathbb{R}^{V \times V},
\end{equation}
where the diagonal elements $\{ \lambda_i \}$ comprise a set of coefficients. 
\begin{figure}[t]
  \centering
  \small
  \begin{overpic}[width=1.0\linewidth]{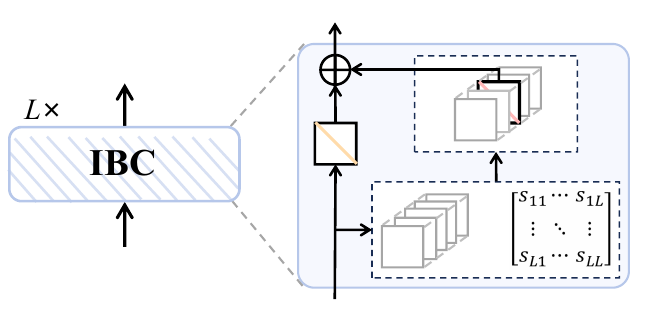}
  \put(72,13){$\times_3$}
\put(59,10){$\mathcal{T}$}
\put(69.8,29.5){$\mathcal{T}_w$}
\put(79.8,27.5){$\Lambda^{\text{LM}}_{\ell}$}
\put(55.3,25.8){$\Lambda^{\text{MC}}_{\ell}$}
  \end{overpic}
  \caption{\textbf{The detailed structure of inter-block communication (IBC) module.} We introduce two coefficient sets, $\Lambda^{\text{MC}}_{\ell}$ and $\Lambda^{\text{LM}}_{\ell}$, the former is communicated under the limitation of HMC, and the latter communicates with other coefficient sets among different blocks. (Best viewed in color).}
  \label{fig_cof}
\end{figure}
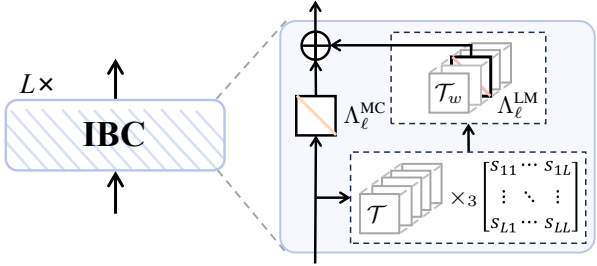

After that, we propose to introduce a learnable relation matrix for achieving inter-block communication. To do this, we need to conceptualize all coefficient set matrices in the entire parameter space as a single tensor. Denote the number of transformer blocks in SAM's image encoder as $L$, and each block contains a PEFT module with a coefficient set that we proposed. We treat the diagonal matrix of each coefficient set as an individual slice in this tensor $\mathcal{T}$, which is derived as:
\begin{equation}
    \mathcal{T} = [\mathbf{\Lambda}_1, \mathbf{\Lambda}_2, \cdots, \mathbf{\Lambda}_L] \in \mathbb{R}^{V \times V \times L}.
\end{equation}
Then, according to the characteristics of gradient propagation in deep learning theory, i.e., chain rule, each frontal slice $\mathbf{\Lambda}_i \in \mathbb{R}^{V \times V}$ of the tensor $\mathcal{T} \in \mathbb{R}^{V \times V \times L}$ is updated sequentially, and thus update the tensor $\mathcal{T}$ is often slow. To avoid the cross-frontal-slice information loss in the tensor $\mathcal{T}$ during learning, we introduce the idea of a special tensor product, i.e., 
\textbf{T-product}.

\noindent \textbf{Definition 4.1. (T-product)} For $\mathcal{A} \in \mathbb{R}^{n_1 \times n_2 \times n_3}$ and $\mathcal{B} \in \mathbb{R}^{n_2 \times l \times n_3}$, the T-product $\mathcal{C} \in \mathbb{R}^{n_1 \times l \times n_3} = \mathcal{A} * \mathcal{B}$ is defined as:
\begin{equation} \label{e.q 7}
    \mathcal{C} = \mathcal{A} * \mathcal{B} = \mathtt{fold}(\mathtt{bcirc}(\mathcal{A}) \cdot \mathtt{unfold}(\mathcal{B})),
\end{equation}
where 
\begin{gather}
    \mathtt{bcric}(\mathcal{A}) = \begin{bmatrix}
        \mathbf{A}^{(1)} & \mathbf{A}^{(n_3)} & \cdots & \mathbf{A}^{(2)} \label{e.q 8}\\
        \mathbf{A}^{(2)} & \mathbf{A}^{(1)}   & \cdots & \mathbf{A}^{(3)} \\
        \vdots           &     \vdots         & \ddots &   \vdots         \\
        \mathbf{A}^{(n_3)} & \mathbf{A}^{(n_3 -1)} & \cdots & \mathbf{A}^{(1)}
    \end{bmatrix}, \\
    \mathtt{unfold}(\mathcal{A}) = [\mathbf{A}^{(1)}, \mathbf{A}^{(2)}, \cdots, \mathbf{A}^{(n_3)}]^T, \\
    \mathtt{fold}(\mathtt{unfold}(\mathcal{A})) = \mathcal{A},
\end{gather}
$\mathbf{A}^{(i)}$ denotes the $i$-th frontal slice $\mathcal{A}(:,:,i)$ of $\mathcal{A}$. There is an invertible linear transform $S: \mathbb{R}^{n_1 \times n_2 \times n_3} \rightarrow \mathbb{R}^{n_1 \times n_2 \times n_3}$ and it transforms the Eq.~\eqref{e.q 7} as
\begin{equation} \label{e.q 11}
    \mathcal{C} = \mathtt{S}^{-1}(S(\mathcal{A}) \odot \mathtt{S}(\mathcal{B}) ) = \mathtt{S}^{-1}(\bar{\mathcal{A}} \odot \bar{\mathcal{B}} )= \mathtt{S}^{-1}(\bar{\mathcal{C}}),
\end{equation}
where $\bar{\mathcal{C}} = \bar{\mathcal{A}} \odot \bar{\mathcal{B}}$ denotes the frontal-slice-wise product (Definition 2.1 refers to~\cite{TTP_2015_LAA}) $\bar{\mathbf{C}}^{(i)} = \bar{\mathbf{A}}^{(i)}  \bar{\mathbf{B}}^{(i)}, i = 1,2,\cdots, n_3$. According to the definition of the frontal-slice-wise product, the invertible linear transform $S$ is formulated as:
\begin{equation} \label{e.q 12}
    \Bar{\mathcal{A}} = \mathtt{S}(\mathcal{A}) =  \mathcal{A} \times_3 \mathbf{S},
\end{equation}
where ``$\times_3$'' denotes the mode-$3$ product and $\mathbf{S} \in \mathbb{R}^{n_3 \times n_3}$ is an arbitrary invertible matrix. Similarly, the inverse transform of Eq.~\eqref{e.q 12} is derived as:
\begin{equation} \label{e.q 13}
    \mathcal{A} = \mathtt{S}^{-1}(\bar{\mathcal{A}}) =  \bar{\mathcal{A}} \times_3 \mathbf{S}^{-1}.
\end{equation}

\noindent \textit{Derivation.} please refer to supplementary material.

\hfill $\blacksquare$

According to Eqs. \eqref{e.q 11}, \eqref{e.q 12}, and \eqref{e.q 13}, we adopt its idea and
design an arbitrary invertible relation matrix $\mathbf{S} \in \mathbb{R}^{L \times L}$ to capture the cross-slice information in $\mathcal{T}$. Then the whole tensor $\mathcal{T}_w$ is formulated as: 
\begin{align}
\label{e.q 4}
    \mathcal{T}_w &= \mathcal{T}\times_3 \mathbf{S} \nonumber  \\
                  &= [\mathbf{\Lambda}^{l}_{1}, \mathbf{\Lambda}^{l}_{2}, \cdots, \mathbf{\Lambda}^{l}_{L}] \in \mathbb{R}^{V \times V \times L}, 
\end{align}
where $\times_3$ denotes mode-$3$ product and the relation matrix $\mathbf{S}$ is learnable. 


\noindent\textbf{Dual Coefficient Sets.} Although HMC limits the substantial adjustment of parameter space, it preserves as much of the task-relevant information as possible when propagating through sequential layers~\cite{information_bottleneck_2019_IOP,information_2017_arxiv}.
In order to retain this advantage, as shown in Fig.~\ref{fig_cof}, we further extend a single coefficient set to dual coefficient sets: one set $\{ \mathbf{\Lambda}^{\text{MC}}_{\ell}\}$ communicates under the constraint of HMC, while the other $\{ \mathbf{\Lambda}^{\text{LM}}_{\ell}\}$ communicates among different layers.


\subsection{Intra-block Enhancement}
\label{hcl}

Since the relation matrix applies a uniform weight distribution across all coefficient elements in the same block, it inherently lacks the ability to apply distinct adjustments for individual elements. To address this limitation, we introduce an intra-block enhancement module that involves a hyper-complex layer (HL) to generate weights $\textbf{W}$ for a linear projection head parameterized.

\begin{figure}[t]
  \centering
  \small
  \begin{overpic}[width=1.0\linewidth]{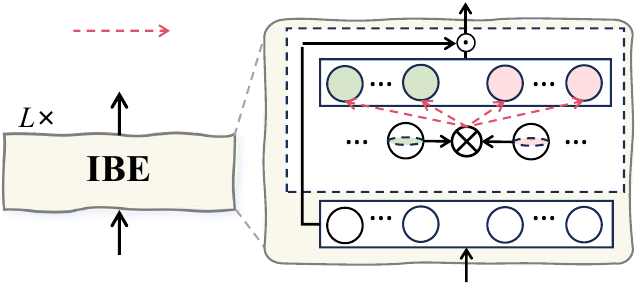}
  \put(7.6,37.6){\normalsize{$\mathbb{H}$}}
  \put(28,37.6){\normalsize{$\mathbb{R}$}}
  \put(15,41.6){\footnotesize{Proj}}
  \put(52,16.6){\large{HL}}
  \end{overpic}
  \caption{\textbf{The detailed structure of intra-block enhancement (IBE) module.} We introduce a hyper-complex layer (HL) for facilitating communication among projection directions in each layer. ``Proj'': Projection. ``HL'': hyper-complex layer. $\mathbb{H}$: hyper-complex space, i.e., suprasphere. (Best viewed in color) ``$\otimes$'': $\mathtt{Hamilton}$ $\mathtt{product}$.}
  \label{fig_hc}
\end{figure}
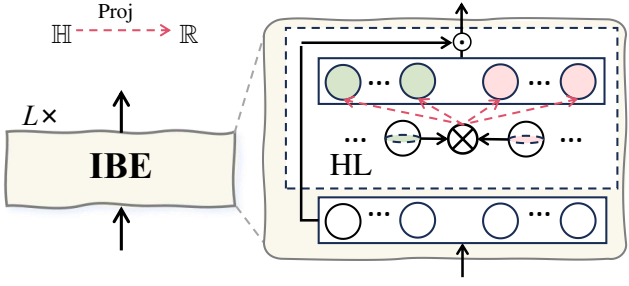


Specifically, in HL, the weights are 
obtained from a suprasphere, which is initialized as orthogonal weights, as shown in Fig.~\ref{fig_hc}.
Then, we use the $ \mathtt{ Hamilton}$ $\mathtt{product}$  $\otimes$~\cite{NN_1994_IEEE} to update the element $H$ of the suprasphere. Define two weights of a element $\widetilde{H_{a}}$ and $\widetilde{H_{b}}$ respectively as follows:
\begin{align}
    \widetilde{H_{a}} = a_01 + a_1j_1 + \cdots  + a_{N-1}j_{N-1} \\
    \widetilde{H_{b}} =  b_01 + b_1j_1  + \cdots  + b_{N-1}j_{N-1}.
\end{align}
Then, we obtain the corresponding updated element $W_{i}$ via a hyper-complex layer, which is formulated as: 
\begin{align} 
    H =& \widetilde{H_{a}} \otimes  \widetilde{H_{b}} \nonumber \\
    =& (a_0b_0 + \cdots + a_0b_{N-1}j_{N-1})1 + \nonumber \\
    & (a_1b_0  + \cdots + a_1b_{N-1}j_{N-1})j_1 + \nonumber \\
    & \qquad \qquad\qquad \cdots \nonumber  \\
    & (a_{N-1}b_0 +  \cdots + a_{N-1}b_{N-1}j_{N-1})j_{N-1}. \label{e.q 18}
\end{align}
In the HL, all parameters are hyper-complex numbers, including elements and weights. 
The $ \mathtt{ Hamilton}$ $\mathtt{product}$ performs transformations of elements in hyper-complex space, as well as scaling and interpolation between two rotations following a geodesic over a sphere in the $\mathbb{R}^{N-1}$ suprashpere. More details of the forward propagation, the back-propagation, and the parametrization of the hyper-complex layer can be found in supplementary material. Finally, HL uses a real transform $Q: \mathbb{H}^{N} \rightarrow \mathbb{R}^{V}$ to transform the suprasphere back to the parameter space, and then obtain the corresponding parameter weights $\textbf{W}$. This is usually achieved by multiple concatenating elements in the suprasphere via a specific rule~\cite{hyper_number_1972_science}. By equipping the projection head with HL, the adjustment of individual coefficient elements can be enhanced to achieve a greater impact on the projection direction of the parameter space.


\subsection{Overall Architecture}
Overall, we develop a SAM-COBOT framework, and for a specific input feature map $\textbf{M}_{\ell}$ in the ${\ell}^{\text{th}}$ SAM-COBOT module, the right branch in the SAM-COBOT module produces the adjusted feature map, $\widetilde{\textbf{M}}_{\ell}$, formally via:

\begin{equation}
  \widetilde{\textbf{M}}_{\ell} = \mathcal{F}_{\ell}(\textbf{M}_{\ell}; \textbf{W}  \mathbf{\Lambda}^{\text{MC}}) + \mathcal{F}_{\ell}(\textbf{M}_{\ell}; \textbf{W}  \mathbf{\Lambda}^{\text{LM}}),
\end{equation}
where $\mathcal{F}_{\ell}$ represents ${\ell}^{\text{th}}$ block of SAM's image encoder.



\noindent\textbf{Fine-tuning.} During the fine-tuning phase, SAM-COBOT is fine-tuned in conjunction with the existing PEFT modules. Concurrently, the original components of SAM load their weights from the pre-trained checkpoint, with their parameters remaining frozen.

\noindent\textbf{Loss Function.} Following previous works~\cite{SonarSAM_arxiv_2023,SAM_PEFT_MedSAM_arxiv_2023}, we incorporate a combination of binary cross-entropy loss, denoted as $\mathcal{L}_{\text{ce}}$, and binary dice loss, represented by $\mathcal{L}_{\text{dice}}$, for the fine-tuning of SAM. The overall loss function is derived as:
\begin{equation}
\mathcal{L} = \mathcal{L}_{\text{ce}} + \mathcal{L}_{\text{dice}}
\end{equation}

\section{Experiments}

In this section, we evaluate our SAM-COBOT on a diverse range of downstream segmentation tasks. These tasks can be broadly classified into three main categories: (1) Medical image segmentation, (2) Natural image segmentation, and (3) Remote sensing image segmentation. We begin with a description of the datasets used, followed by the associated evaluation metrics, baseline models, and implementation details. Subsequently, we conduct an ablation study to evaluate the individual contributions of components in our proposed SAM-COBOT. Finally, we compare SAM-COBOT with other prevalent parameter-efficient fine-tuning (PEFT) techniques.

\subsection{Experimental Setup}
\noindent\textbf{Dataset.} We evaluate the performance of our method on 10 datasets. These datasets cover multiple tasks of natural image segmentation (COCO~\cite{Data_MSCOCO_2014_ECCV}, TRCAN~\cite{DATA_trashcan_2020_arxiv}), remote sensing image segmentation (NWPU~\cite{Data_NWPU3_2016_TGRS,Data_NWPU1_2014_ISPRS,Data_NWPU2_2016_ISPRS}, SSDD~\cite{Data_SSDD_RS_2021}, SONAR~\cite{DATA_SONAR_2021_ICCVW}), and medical image segmentation (ADOME~\cite{Data_AdomenCT_1K_Tpami_2022}, SPLEN~\cite{DATA_MULTI_TASK_2022_NATURE_C}, MOMO~\cite{DATA_MULTI_TASK_2022_NATURE_C}, BRAST~\cite{DATA_BREAST_2020_BRIEF}, SEGRAP~\cite{DATA_segrap_2023_comp}) .



\noindent\textbf{Evaluation Metrics.} In line with previous studies~\cite{SAM_PEFT_MedSAM_arxiv_2023,SonarSAM_arxiv_2023}, we utilize the Dice Similarity Coefficient (DSC) for evaluating medical image segmentation. For both natural and remote sensing image segmentation, we adopt mean intersection-over-union (mIoU).

\noindent\textbf{Baseline Models.} We implement our SAM-COBOT onto two popular PEFT
methods for fine-tuning SAM~\cite{FM_SAM_2023_ICCV}, i.e., LoRA~\cite{PEFT_LORA_2022_ICLR} and Adaptformer~\cite{PEFT_adaptformer_2022_nips}.

\noindent\textbf{Implementation Details.} 
In all of our experiments, we employ the ViT-Base version of SAM~\cite{FM_SAM_2023_ICCV} as our backbone, integrating a box prompt for its prompt encoder input. In line with previous studies~\cite{SAM_PEFT_MedSAM_arxiv_2023,SonarSAM_arxiv_2023}, we apply a random perturbation to each bounding box, varying between 0 and 50 pixels. Our training employs the Adam optimizer~\cite{adam_arxiv_2014}. 
For medical image segmentation, the initial learning rate is set to $1.25 \times 10^{-6}$, and the weight decay is $5 \times 10^{-4}$ with one image per mini-batch. The number of fine-tuning epochs is set to 25. For natural and remote sensing image segmentation, we follow SonarSAM~\cite{SonarSAM_arxiv_2023}, the initial learning rate is set to $10^{-4}$, and the weight decay is $5 \times 10^{-5}$ with one image per mini-batch. The number of fine-tuning epochs is set to 20. More details are provided in the supplementary material.


\begin{table}[t]
\small
\tabcolsep=0.05cm
\begin{center}
\renewcommand\arraystretch{1.1}
\setlength{\tabcolsep}{6pt}{
\begin{tabular}{ c  c  c | c | c | c}
\toprule
CoS & RM & HL &ADOME & NWPU & TRCAN \\
\midrule 
\xmark & \xmark & \xmark & 90.1 & 83.0 & 73.3  \\
\cmark & \xmark & \xmark & 90.1 \scriptsize{\color{Highlight}{($+$0.0})} & 83.1 \scriptsize{\color{Highlight}{($+$0.1})} & 73.4 \scriptsize{\color{Highlight}{($+$0.1})}  \\
\xmark & \xmark & \cmark & 90.3 \scriptsize{\color{Highlight}{($+$0.2})} & 83.3 \scriptsize{\color{Highlight}{($+$0.3})} & 73.5 \scriptsize{\color{Highlight}{($+$0.2})}  \\
\cmark & \xmark & \cmark & 90.4 \scriptsize{\color{Highlight}{($+$0.3})} & 83.3 \scriptsize{\color{Highlight}{($+$0.3})} & 73.5 \scriptsize{\color{Highlight}{($+$0.2})}  \\
\cmark & \cmark & \xmark & 90.9 \scriptsize{\color{Highlight}{($+$0.8})} & 83.7  \scriptsize{\color{Highlight}{($+$0.7)}} & 73.9 \scriptsize{\color{Highlight}{($+$0.6})}  \\
\cmark & \cmark & \cmark & \CC{15}91.3 \scriptsize{\color{Highlight}{($+$1.2})} & \CC{15}84.0  \scriptsize{\color{Highlight}{($+$1.0})} & \CC{15}74.1 \scriptsize{\color{Highlight} {($+$0.8})}   \\ \bottomrule
\end{tabular}}
\vspace{1mm}
\caption{\textbf{Ablation study results (\%) on three datasets}: ADOME~\cite{Data_AdomenCT_1K_Tpami_2022}, NWPU~\cite{Data_NWPU3_2016_TGRS,Data_NWPU1_2014_ISPRS,Data_NWPU2_2016_ISPRS}, and TRCAN~\cite{DATA_trashcan_2020_arxiv}. The baseline is Adaptformer~\cite{PEFT_adaptformer_2022_nips}.
Their results are shown in the first row. ``CoS'': dual coefficient sets.
``RM'': relation matrix, and ``HL'': hypercomplex layer.}
\label{tab_ablation}
\end{center}
\vspace{-5mm}
\end{table}

    

\begin{table}[t]
\small
\tabcolsep=0.05cm
\begin{center}
\renewcommand\arraystretch{1.1}
\setlength{\tabcolsep}{6pt}{
\begin{tabular}{ c | c | c | c}
\toprule
Strategy of RM &ADOME & NWPU & TRCAN \\
\midrule 
Fixed  &90.2 ± 0.6 & 83.4 ± 0.6 & 73.3 ± 0.2  \\
Learnable  & \CC{15}91.3 ± 0.5 & \CC{15}84.0 ± 0.3 & \CC{15}74.1 ± 0.0  \\ \bottomrule 

\end{tabular}}
\vspace{1mm}
\caption{\textbf{Effects of RM on three datasets}: ADOME~\cite{Data_AdomenCT_1K_Tpami_2022}, NWPU~\cite{Data_NWPU3_2016_TGRS,Data_NWPU1_2014_ISPRS,Data_NWPU2_2016_ISPRS}, and TRCAN~\cite{DATA_trashcan_2020_arxiv}. The baseline model is Adaptformer~\cite{PEFT_adaptformer_2022_nips}. ``Fixed'': random values. ``Learnable'': update by back-propagation (i.e., ours).}
\label{tab_dcm}
\end{center}
\end{table}

\begin{table}
\small
\vspace{-2mm}
\tabcolsep=0.05cm
\begin{center}
\renewcommand\arraystretch{1.1}
\setlength{\tabcolsep}{6pt}{
\begin{tabular}{ c c | c | c | c}
\toprule
Linear & HL &ADOME & NWPU & TRCAN \\
\midrule 
\cmark & \xmark & 90.9 ± 0.6 & 83.8 ± 0.2 & 73.9 ± 0.1 \\
\xmark & \cmark  & \CC{15}91.3 ± 0.5 & \CC{15}84.0 ± 0.3 & \CC{15}74.1 ± 0.0 \\ \bottomrule

\end{tabular}}
\vspace{1mm}
\caption{\textbf{Discussion of hyper-complex layer on three datasets}: ADOME~\cite{Data_AdomenCT_1K_Tpami_2022}, NWPU~\cite{Data_NWPU3_2016_TGRS,Data_NWPU1_2014_ISPRS,Data_NWPU2_2016_ISPRS}, and TRCAN~\cite{DATA_trashcan_2020_arxiv}. The baseline model is Adaptformer~\cite{PEFT_adaptformer_2022_nips}. ``Linear'': a linear layer.}
\label{tab_dHL}
\end{center}
\vspace{-5mm}
\end{table}

\subsection{Ablative Studies}


\noindent\textbf{Ablation of Main Components.}
Here, we do an ablation study to show the benefit brought by each component of our proposed SAM-COBOT, i.e., coefficient set (CoS), relation matrix (RM), and hyper-complex layer (HL) on three datasets, including ADOME~\cite{Data_AdomenCT_1K_Tpami_2022}, NWPU~\cite{Data_NWPU3_2016_TGRS,Data_NWPU1_2014_ISPRS,Data_NWPU2_2016_ISPRS} and TRCAN datasets~\cite{DATA_trashcan_2020_arxiv}. We use Adaptformer~\cite{PEFT_adaptformer_2022_nips} as the baseline in row 1 of Table ~\ref{tab_ablation}. Comparing row 2 to row 1, we can
see slight performance gains brought by the coefficient set, as it introduces more parameter space to be optimized. Then, solely introducing the hyper-complex layer shows limited improvement as it can only adjust projection directions within each layer. Moreover, the results are further boosted by large margins after introducing the relation matrix, showing its capability to capture interdependencies among different layers. Finally, by integrating the hyper-complex layer, the results reveal clear performance gains, e.g., 1.2\% on ADOME dataset. 


\begin{table*}
\centering
\small
\tabcolsep=0.032cm
\begin{tabular}{l|c|cccccccccc|c}

\toprule
\multirow{2}{*}{Method} & \multirow{2}{*}{Params(K)} & \multicolumn{2}{c}{\textbf{Natural}} & \multicolumn{3}{c}{\textbf{Remote Sensing}} & \multicolumn{5}{c|}{\textbf{Medical}} & \multirow{2}{*}{\textbf{Avg}}\\
\cmidrule(r){3-4} \cmidrule(r){5-7} \cmidrule(r){8-12}
& & COCO & TRCAN & NWPU & SSDD & SONAR & ADOME & SPLEN & MOMO & BRAST & SEGRAP \\
\midrule
{\color{Gray} Freeze} &  0 &{\color{Gray} 53.0 \scriptsize{± 0.1}} & {\color{Gray} 53.9 \scriptsize{± 0.2}} & {\color{Gray} 59.6 \scriptsize{± 0.9}} & {\color{Gray} 63.2 \scriptsize{± 0.2}} &{\color{Gray} 34.5 \scriptsize{± 2.7}}  & {\color{Gray} 23.5 \scriptsize{± 1.2}} & {\color{Gray} 24.3 \scriptsize{± 11.5}} & {\color{Gray} 24.3 \scriptsize{± 3.3}} & {\color{Gray} 60.1 \scriptsize{± 1.4}} & {\color{Gray} 10.5 \scriptsize{± 0.2}} & 40.7 \\

Lightweight & 0 & 70.4 \scriptsize{± 0.1}  & 70.3 \scriptsize{± 0.2} & 80.5 \scriptsize{± 0.1} & 80.2 \scriptsize{± 0.2} & 79.8 \scriptsize{± 0.1} & 86.0 \scriptsize{± 0.4} & 93.4 \scriptsize{± 1.2} & 86.3 \scriptsize{± 3.0} & 85.1 \scriptsize{± 0.6} & 67.8 \scriptsize{± 0.2} & 80.0 \\ \midrule

\multicolumn{13}{c}{\textbf{\CC{15} \emph{Parameter-efficient Fine-Tuning}}}
\\

LoRA~\cite{PEFT_LORA_2022_ICLR} & 147.4 & 71.8 \scriptsize{± 0.1} & 72.8 \scriptsize{± 0.1} & 81.8 \scriptsize{± 0.2} & 80.7 \scriptsize{± 0.1} & 82.8 \scriptsize{± 0.1} & 88.0 \scriptsize{± 0.4}  & 94.4 \scriptsize{± 0.4} & 86.6 \scriptsize{± 2.2} & 84.8 \scriptsize{± 0.6} & 68.7 \scriptsize{± 0.1} & 81.1\\

LoRA~\cite{PEFT_LORA_2022_ICLR}+Ours & 148.3 \color{frenchblue}\scriptsize{($+$0.9)}& \textbf{72.1} \scriptsize{± 0.1} & \textbf{73.1} \scriptsize{± 0.0} & \textbf{82.5} \scriptsize{± 0.2}  & \textbf{81.2} \scriptsize{± 0.1} & \textbf{84.6} \scriptsize{± 0.1}  & \textbf{88.7} \scriptsize{± 0.2} & \textbf{94.9} \scriptsize{± 0.1} & \textbf{86.7} \scriptsize{± 2.4} & \textbf{85.3} \scriptsize{± 1.2} & \textbf{70.1} \scriptsize{± 0.1} & \textbf{81.8}\\

Adaptformer~\cite{PEFT_adaptformer_2022_nips}  & 322.7 & 71.7 \scriptsize{± 0.1} & 73.3 \scriptsize{± 0.1} & 83.0 \scriptsize{± 0.1} & 81.9 \scriptsize{± 0.1} & 84.1 \scriptsize{± 0.1} & 90.1 \scriptsize{± 0.2} & 94.8 \scriptsize{± 0.5} & 87.6 \scriptsize{± 3.1} & 85.8 \scriptsize{± 0.2} & 72.1 \scriptsize{± 0.1} & 82.4 \\

Adaptformer~\cite{PEFT_adaptformer_2022_nips}+Ours & 324.0 \color{frenchblue}\scriptsize{($+$1.3)} & \textbf{72.2} \scriptsize{± 0.0} & \textbf{74.1} \scriptsize{± 0.0} & \textbf{84.0} \scriptsize{± 0.3}  & \textbf{82.4} \scriptsize{± 0.2} & \textbf{84.9} \scriptsize{± 0.1}  & \textbf{91.3} \scriptsize{± 0.5} & \textbf{96.4} \scriptsize{± 1.7} & \textbf{89.2} \scriptsize{± 1.4} & \textbf{87.3} \scriptsize{± 0.6} & \textbf{73.1} \scriptsize{± 0.1} & \textbf{83.6}\\
\bottomrule

\end{tabular}
\caption{\textbf{Segment anything model (SAM) fine-tuned on a diverse range of downstream segmentation tasks, with the corresponding size of trainable parameters.} All results are based on ViT-Base~\cite{CV_vit_2020_arxiv} backbone, and we ignore SAM's lightweight mask decoder when calculating the learnable parameters. We use DSC (\%) for medical image segmentation, and mIoU (\%) for other tasks as evaluation metrics. ``Freeze'': without any form of fine-tuning. ``Lightweight'':  freezes all the backbone parameters and only tunes SAM's lightweight mask decoder. ``Avg'': average.}
\label{tab:global}
\end{table*}


\begin{figure*}[t]
  \centering
  \begin{subfigure}[b]{0.33\linewidth} 
    \includegraphics[width=\linewidth]{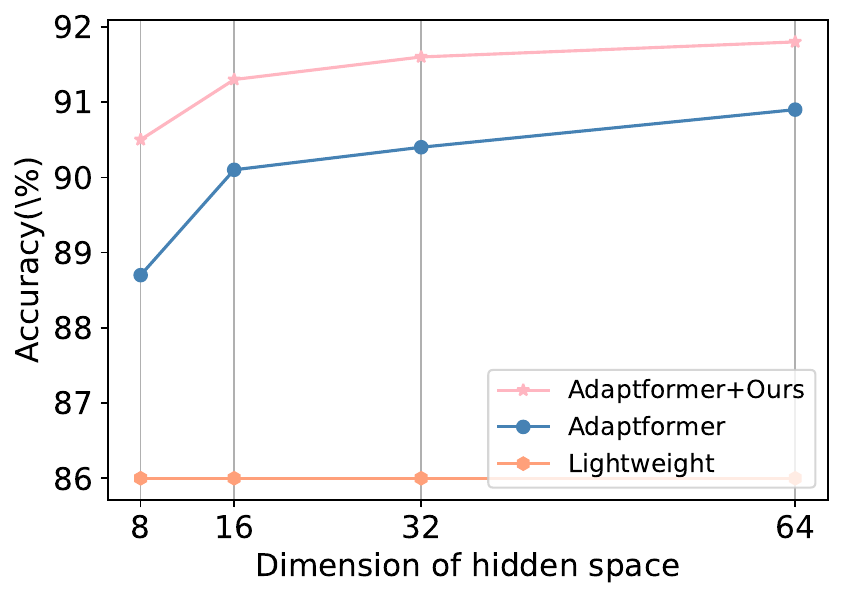}
    \caption{ADOME~\cite{Data_AdomenCT_1K_Tpami_2022}}
  \end{subfigure}
  \hfill 
  \begin{subfigure}[b]{0.33\linewidth}
    \includegraphics[width=\linewidth]{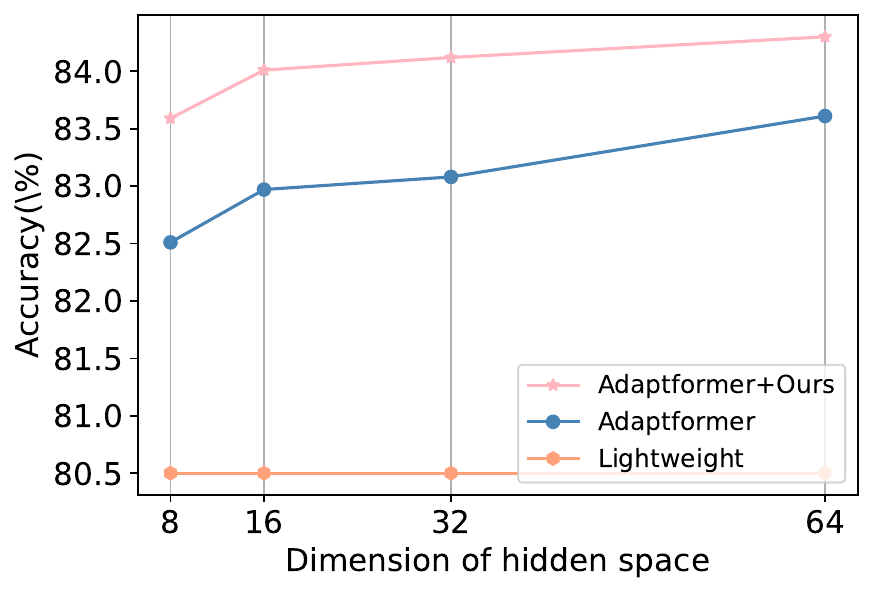}
\caption{NWPU~\cite{Data_NWPU1_2014_ISPRS,Data_NWPU3_2016_TGRS,Data_NWPU2_2016_ISPRS}}
  \end{subfigure}
  \hfill 
  \begin{subfigure}[b]{0.33\linewidth}
    \includegraphics[width=\linewidth]{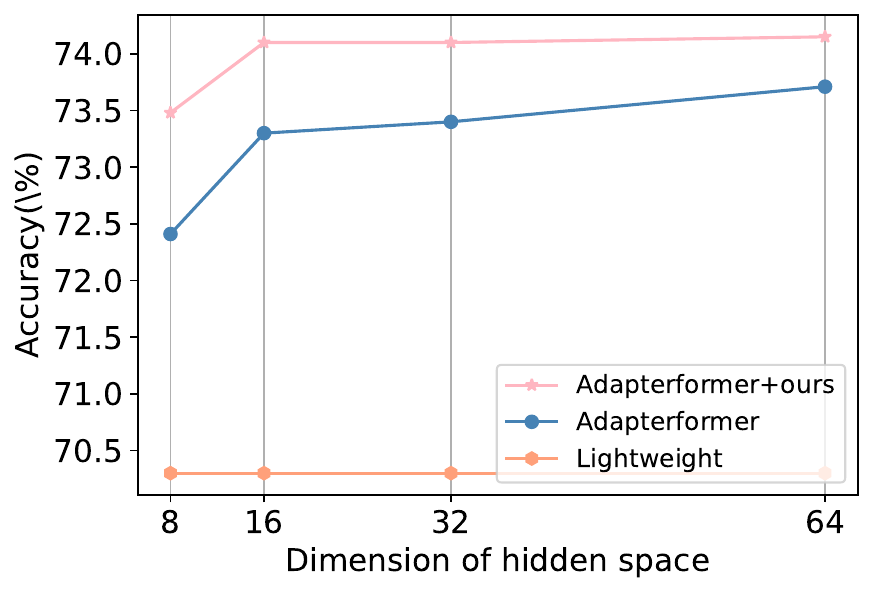}
    \caption{TRCAN~\cite{DATA_trashcan_2020_arxiv}}
  \end{subfigure}
  \caption{\textbf{Results on different dimensions of hidden space $r$}  (Best view in color).}
  \label{fig:dimen}
\end{figure*} 

\noindent\textbf{Effects of Relation Matrix (RM).} In Table~\ref{tab_dcm}, we demonstrate the efficacy of our proposed learnable relation matrix by comparing it with a fixed matrix initialized with random values. We can observe a significant performance improvement with our method, e.g., 1.1\% DSC on ADOME dataset~\cite{Data_AdomenCT_1K_Tpami_2022}. 


\begin{table}[t]
    \centering
    \begin{tabular}{l|cc|cc}
    \toprule
    \multirow{2}*{} & \multicolumn{2}{c|}{\textbf{ViT-Base}} & \multicolumn{2}{c}{\textbf{ViT-Large}} \\ \cline{2-5}
    Method & \rotatebox[origin=c]{90}{SSDD} & \rotatebox[origin=c]{90}{ADMOE} & \rotatebox[origin=c]{90}{SSDD} & \rotatebox[origin=c]{90}{ADMOE} \\ \midrule
    LORA~\cite{PEFT_LORA_2022_ICLR}     & 80.7 & 88.0 & 81.8 & 89.1\\
    LORA~\cite{PEFT_LORA_2022_ICLR}+Ours     &  \textbf{81.2} & \textbf{88.7} & \textbf{82.4} & \textbf{89.9} \\ 
    Adaptformer~\cite{PEFT_adaptformer_2022_nips}     & 81.9 & 90.1 & 82.1 & 91.6 \\
    Adaptformer~\cite{PEFT_adaptformer_2022_nips}+Ours     & \textbf{82.4} & \textbf{91.3} & \textbf{82.8} &  \textbf{93.0} \\ \bottomrule

    \end{tabular} 
    \caption{\textbf{Results on different backbones.} SSDD~\cite{Data_SSDD_RS_2021} and ADMOE~\cite{Data_AdomenCT_1K_Tpami_2022} are two datasets we employed.}
    \label{tab:fb}
\end{table}

\noindent\textbf{Linear Layer or Hyper-complex Layer.} Table~\ref{tab_dHL} presents a comparison between our proposed hyper-complex layer and a standard linear layer, which is a commonly used module for communicating among channels, i.e., projection directions. The results reveal improvements in performance, e.g., 0.4\% DSC ADOME dataset~\cite{Data_AdomenCT_1K_Tpami_2022}. This suggests that the orthogonality facilitated by the hyper-complex layer plays a beneficial role in enhancing intra-layer communication.

\subsection{Main Results}

\textbf{Comparing to SOTA.} We compare our approach against several prevailing PEFT techniques for SAM, including LoRA~\cite{PEFT_LORA_2022_ICLR} and Adapterformer~\cite{PEFT_adaptformer_2022_nips}, on 10 datasets across three domains in the computer vision community. We present their original results and also show our results (by plugging SAM-COBOT in these methods) in Table~\ref{tab:global}. As shown in Table~\ref{tab:global}, our SAM-COBOT becomes the new state-of-the-art. Remarkably, our SAM-COBOT, with minimal parameter overhead, significantly enhances both LoRA and Adaptformer. This improvement is particularly noticeable in medical image segmentation.  By way of illustration, our SAM-COBOT boosts Adaptformer by 1.2\% and 1.0\% in terms of mIoU on ADOME~\cite{Data_AdomenCT_1K_Tpami_2022} and SEGRAP~\cite{DATA_segrap_2023_comp} dataset, respectively. Notably, our method achieves 0.5\% DSC gains on LoRA on BRAST dataset, despite starting from a lower performance (i.e., 84.8\%) than the baseline (i.e., 85.1\%). Notably, LoRA achieves a 1.1\% improvement through lightweight fine-tuning on average  by incorporating 147.4K parameters. In contrast, SAM-COBOT, with a mere addition of 0.9K parameters, achieves an additional 0.7\% enhancement in performance. The above results underscore the robustness and generalization capabilities of our SAM-COBOT. More results are provided in the supplementary material.

\begin{figure*}[t]
  \centering
  \begin{overpic}[width=1.0\textwidth]{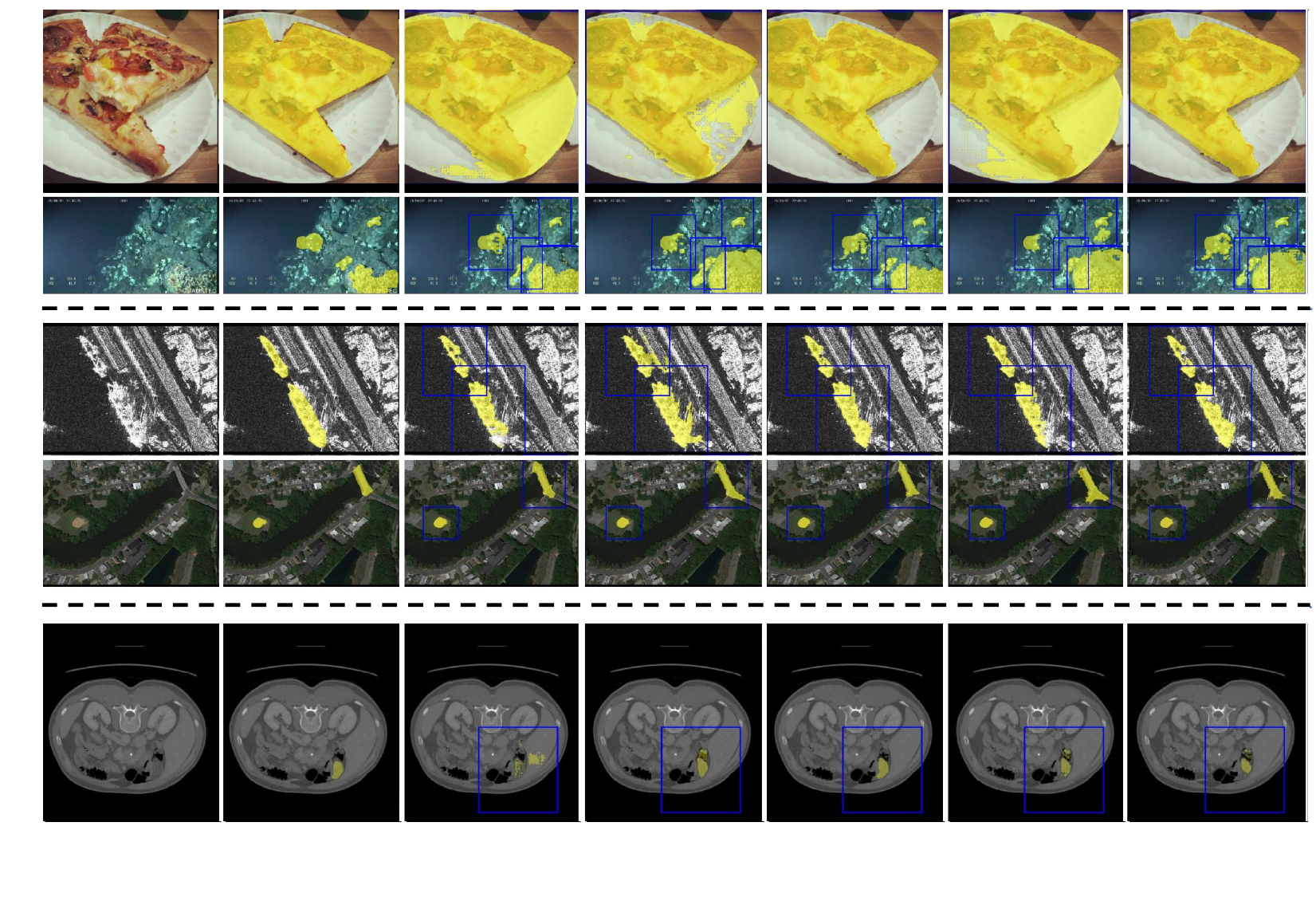}
  \put(0.0,60.5){\large{(a)}}
  \put(0.0,49.3){\large{(b)}}  
  \put(0.0,37.8){\large{(c)}}  
  \put(0.0,28.3){\large{(d)}}  
  \put(0.0,13.2){\large{(e)}} 

  \put(8,3){\normalsize{Input}}
  \put(17.8,3){\normalsize{Ground Truth}}
  \put(32.1,3){\normalsize{Lightweight}}
  \put(45.7,3){\normalsize{Adaptformer}}
  \put(57.5,3){\normalsize{Adaptformer+Ours}}
  \put(76.5,3){\normalsize{LoRA}}
  \put(87.5,3){\normalsize{LoRA+Ours}}

  \end{overpic}
  \vspace{-27pt}
  \caption{\textbf{Qualitative segmentation results on three scenarios}, i.e., (a) natural image segmentation on COCO dataset~\cite{Data_MSCOCO_2014_ECCV}, (b) natural image segmentation on TRCAN~\cite{DATA_trashcan_2020_arxiv} dataset, (c) remote sensing image segmentation on SSDD~\cite{Data_SSDD_RS_2021} dataset, (d) remote sensing image segmentation on NWPU~\cite{Data_NWPU3_2016_TGRS,Data_NWPU1_2014_ISPRS,Data_NWPU2_2016_ISPRS} dataset and (e) medical image segmentation on ADOME~\cite{Data_AdomenCT_1K_Tpami_2022} dataset. ``Lightweight'':  freezes all the backbone parameters and only tunes SAM’s lightweight
mask decoder.}

  \label{fig:vis}
\end{figure*}
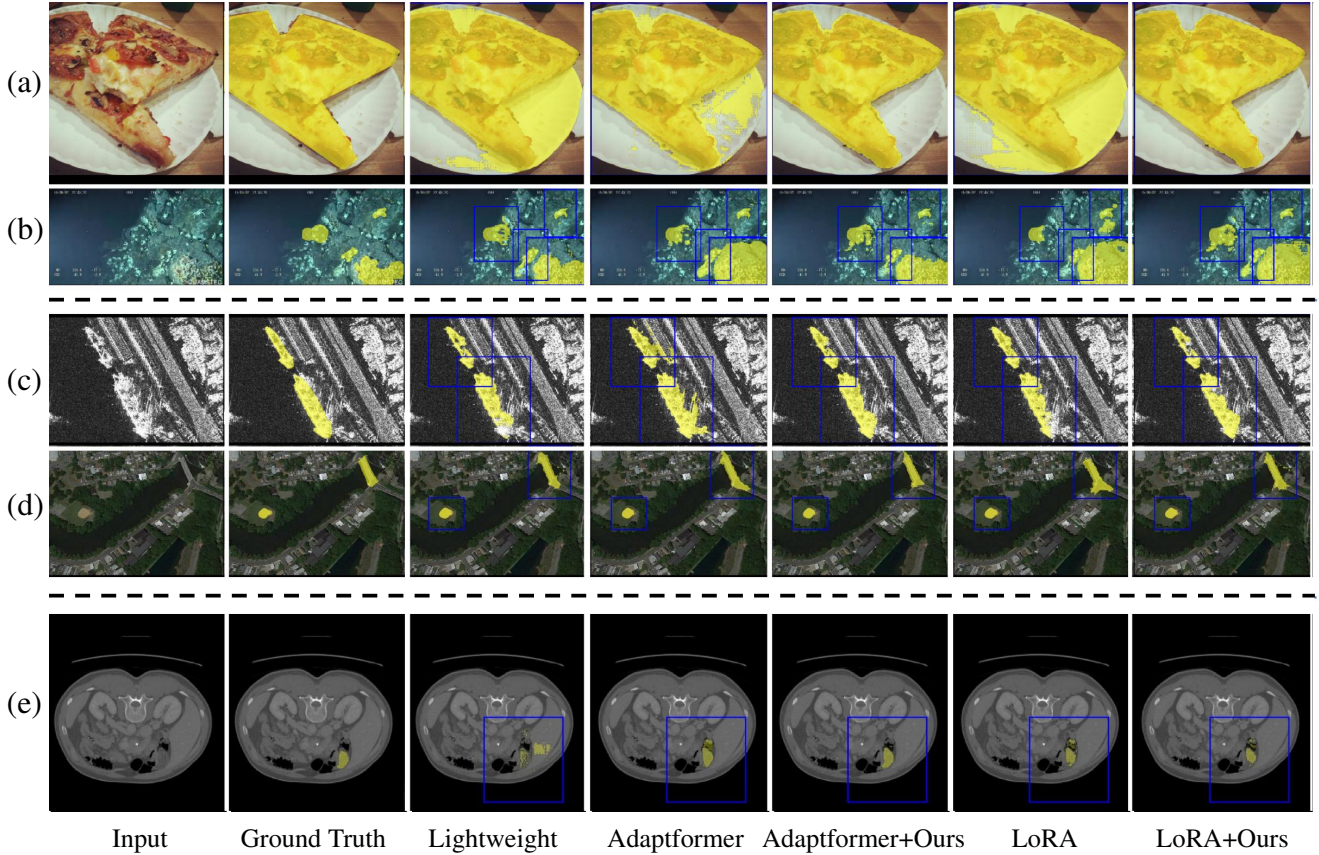

\noindent\textbf{Qualitative results.} Here, we visualize our method's representative example segmentation results against prevailing fine-tuning methods, e.g., LoRA~\cite{PEFT_LORA_2022_ICLR} and Adaptformer~\cite{PEFT_adaptformer_2022_nips} in five datasets. As shown in Fig.~\ref{fig:vis}, we observe that our approach is able to generalize on diverse scenarios and produce more accurate results.

\noindent\textbf{Different Backbones.} We extend our fine-tuning to include larger-scale backbones, e.g., ViT-Base and ViT-Large, reinforcing the versatility of our method. This is evaluated on the SSDD and ADOME datasets, where, as Table~\ref{tab:fb} illustrates, performance improvements are observed consistently. These results demonstrate the generalization of our approach across various transformer architectures.


\noindent\textbf{Different Hidden Dimensions.} In Fig.~\ref{fig:dimen}, we compare our method with a baseline model, specifically Adaptformer, across various dimensions of the hidden space $V$. Overall, our method demonstrates distinct advantages in all dimensional settings. It is noteworthy that at lower dimensions, e.g., $V \leq 16$, our method achieves more pronounced performance improvements, exceeding 1.6\% on the ADOME dataset~\cite{Data_AdomenCT_1K_Tpami_2022}. This observation underscores the efficiency of our method in facilitating interaction among bases, particularly when the number of bases, or \( V \), is constrained.

\section{Conclusion}
In this paper, we equipped PEFT with a cross-block orchestration mechanism to enable the adaptation of the Segment Anything Model (SAM) to various downstream scenarios, called SAM-COBOT. Specifically, SAM-COBOT introduced a novel inter-block communication module to ensure a comprehensive adjustment of the coefficients for each project direction across the entire parameter space, and an intra-block enhancement module to enhance the coordination of projection directions. By incorporating two modules, SAM-COBOT achieved a proper adjustment of parameter space for new scenarios. Extensive experiments showed that the proposed SAM-COBOT can be easily plugged-and-play and consistently improve two prevalent PEFT paradigms by a large margin across three prevalent scenarios, while only introducing 1K additional parameters.

\section{Acknowledgments} This work was supported by NSFC 62322604, 62176159, Natural Science Foundation of Shanghai 21ZR1432200, and Shanghai Municipal Science and Technology Major Project 2021SHZDZX0102.

\clearpage

{
    \small
    \bibliographystyle{ieeenat_fullname}
    \bibliography{main}
}


\end{document}